\documentclass[letterpaper, 10 pt, conference]{ieeeconf}  

\IEEEoverridecommandlockouts                              

\usepackage{cite}
\usepackage{amsmath,amssymb,amsfonts}
\usepackage{algorithmic}
\usepackage{graphicx}
\usepackage{textcomp}
\usepackage{xcolor}

\usepackage[hidelinks]{hyperref}
\usepackage{subcaption}
\usepackage[compatibility=false]{caption}

\title{\LARGE \bf
Early Detection of Visual Impairments at Home \\ Using a Smartphone Red-Eye Reflex Test
}

\author{Judith Massmann$^{1}$, Alexander Lichtenstein$^{2}$, and Francisco M. López$^{3}$
\thanks{The authors thank Dr. Robert D. Williams MD, Gary Franklin, and the KidsVisionCheck contributors for providing the data used in this study. FL acknowledges support from ``The Adaptive Mind'' funded by the Hessian Ministry of Higher Education, Research, Science and the Arts, Germany.}
\thanks{$^{1}$Judith Massmann is with Health Access LLC, 
Friday Harbor, Washington, USA.
        {\tt\small judith@kidsvisioncheck.com}}
\thanks{$^{2}$Alexander Lichtenstein is with Health Access LLC, 
Friday Harbor, Washington, USA.
        {\tt\small alex@kidsvisioncheck.com} }%
\thanks{$^{3}$Francisco M. López with Frankfurt Institute for Advanced Studies, 
Frankfurt am Main, Germany.
        {\tt\small lopez@fias.uni-frankfurt.de}}%
}

\begin{document}

\maketitle
\thispagestyle{empty}
\pagestyle{empty}

\begin{abstract}

Numerous visual impairments can be detected in red-eye reflex images from young children. The so-called Bruckner test is traditionally performed by ophthalmologists in clinical settings. 
Thanks to the recent technological advances in smartphones and artificial intelligence, it is now possible to recreate the Bruckner test using a mobile device. In this paper, we present a first study conducted during the development of KidsVisionCheck, a free application that can perform vision screening with a mobile device using red-eye reflex images. The underlying model relies on deep neural networks trained on children's pupil images collected and labeled by an ophthalmologist. With an accuracy of 90\% on unseen test data, our model provides highly reliable performance without the necessity of specialist equipment. 
Furthermore, we can identify the optimal conditions for data collection, which can in turn be used to provide immediate feedback to the users. In summary, this work marks a first step toward accessible pediatric vision screenings and early intervention for vision abnormalities worldwide.

\end{abstract}

\section{Introduction}

Numerous visual impairments, including amblyopia, congenital glaucoma, and retinoblastoma, can be detected in red-eye reflex images. This so-called Bruckner test consists of flashing a light into a patient's eyes and analyzing the properties of the reflection \cite{roe1984light}. In healthy patients, the retina reflects back as a bright, red, evenly distributed disk. Deviations from this expected outcome are typically indicative of a visual impairment. By way of example, asymmetries between the left and right reflexes are commonly associated with amblyopia or strabismus \cite{wright2013pediatric}, a diffuse reflection may be a sign of congenital glaucoma \cite{barke2022pediatric}, and a white reflex can reveal a retinoblastoma \cite{shields2001pediatric}. Importantly, early detection of these impairments plays a crucial role in the course of action for possible interventions \cite{gilbert2001childhood}. However, many parents are unaware of the importance of this test or are unable to grant their children access to a specialist. This is particularly a problem in areas with low resources. As a consequence, nearly \(20\%\) of the children worldwide are at risk of developing childhood blindness \cite{gilbert2001childhood}.

Luckily, two major technological developments from recent years can help address this problem. First of all, deep neural networks trained for image classification achieve super-human accuracy levels \cite{lecun2015deep}. These advances have been particularly relevant in the medical field: artificial models are capable of detecting numerous diseases in different classes of datasets \cite{esteva2017dermatologist}, including images of the retina \cite{thanki2023deep,sengar2023eyedeep,perdomo2019classification,rajalakshmi2018automated}. These technologies have also been successfully applied to red-eye reflex tests \cite{bernard2022eyescreen,da2018segmentation,linde2023automatic}. It should be noted, however, that in these studies the image datasets are typically collected with high-quality equipment and the models are developed to help professionals in their diagnoses, rather than providing users directly with a tool for self-screening.

\begin{figure}[!t]
    \centering
    \begin{subfigure}[t]{0.238\textwidth}
        \centering
        \includegraphics[width=\textwidth]{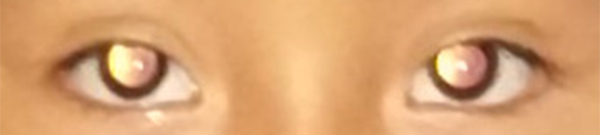}
    \end{subfigure}
    \hfill
    \begin{subfigure}[t]{0.238\textwidth}
        \centering
        \includegraphics[width=\textwidth]{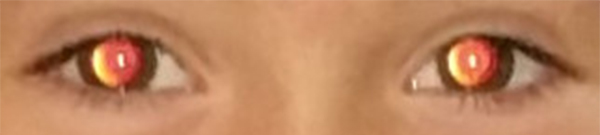}
    \end{subfigure}
    \\[.4em]
    \begin{subfigure}[t]{0.238\textwidth}
        \centering
        \includegraphics[width=\textwidth]{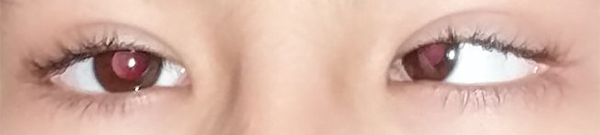}
    \end{subfigure}
    \hfill
    \begin{subfigure}[t]{0.238\textwidth}
        \centering
        \includegraphics[width=\textwidth]{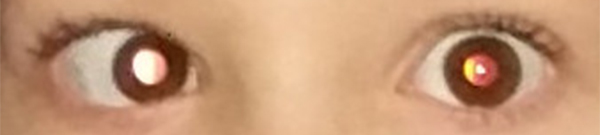}
    \end{subfigure}
    \\[.4em]
    \begin{subfigure}[t]{0.238\textwidth}
        \centering
        \includegraphics[width=\textwidth]{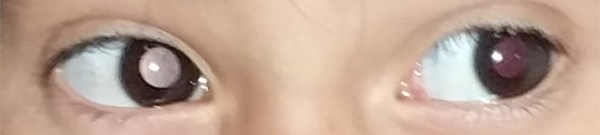}
    \end{subfigure}
    \hfill
    \begin{subfigure}[t]{0.238\textwidth}
        \centering
        \includegraphics[width=\textwidth]{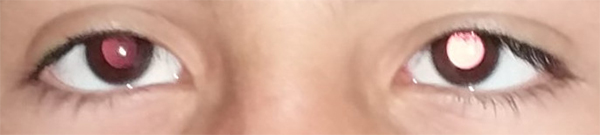}
    \end{subfigure}
    \caption{Examples of visual impairments detected in red-eye reflex images collected for this study.}
    \label{fig:examples}
    \vspace*{-.2cm}
\end{figure}

\begin{figure*}[!t]
    \centering
    \hfill
    \includegraphics[width=.99\textwidth]{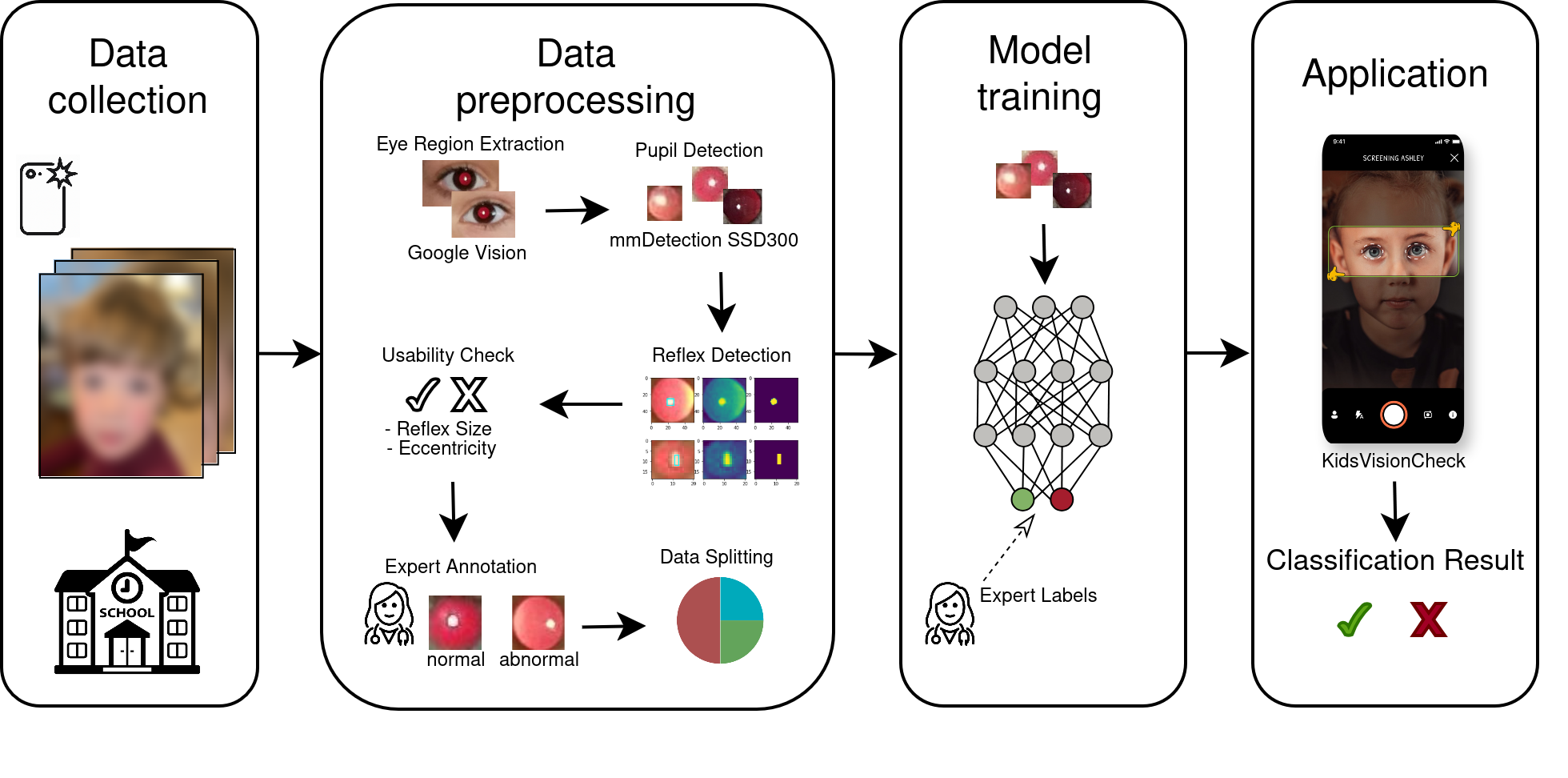}
    \caption{Procedure followed in this study. Images (here blurred for privacy) were collected with a smartphone in schools. During preprocessing, individual pupils were cropped and categorized as useful or not. The useful images were annotated by an ophthalmologist as normal and abnormal and subsequently split into training, testing, and validation datasets. Pre-trained neural networks were used to classify the training images. After training, the model parameters were frozen for testing.}
    \label{fig:KVC_pipeline}
\end{figure*}

That is why the second major technological development, the smartphone, can play such an important role. Mobile devices include good quality cameras, robust computational power, and mainly an internet connection to servers where data and models can provide instantaneous feedback and screening results. For this reason, health evaluations through smartphone apps have recently become popular as of late \cite{anikwe2022mobile}. Circumventing specialists may spark some caution alarms. Nevertheless, it is clear that smartphones can provide access to some form of health evaluations in areas that may lack other alternatives \cite{steinhubl2015emerging,esteva2017dermatologist}. Some recent studies have aimed to implement the Bruckner test in images from mobile devices with both positive \cite{srivastava2022reliability} and negative \cite{tan2025using} outcomes, but with limited applicability beyond the studies themselves.

We put forward a novel implementation of the Bruckner test for smartphones, which will serve as the backbone for the KidsVisionCheck app\footnote{The app includes a preliminary version of this model. More information about it can be found at \href{https://kidsvisioncheck.com/}{https://kidsvisioncheck.com/}}. Our approach combines image analysis techniques with deep neural networks and the evaluation of expert ophthalmologists to design a model that can not only achieve an accuracy of \(90\%\), but also provide feedback to the user about which conditions might enhance the confidence of the models in the classification. The rest of the article is divided as follows: in Section \ref{sec:methodology} we present the methodology followed to collect the data and train the models; in Section \ref{sec:results} we analyze the results and aim to provide explainability; and in Section \ref{sec:discussion} we discuss the findings and limitations and provide an overview of future directions for this research, including a new iteration of data collection based on the outcomes of the present study.

\begin{figure*}[!t]
    \centering
    \includegraphics[width=0.99\textwidth]{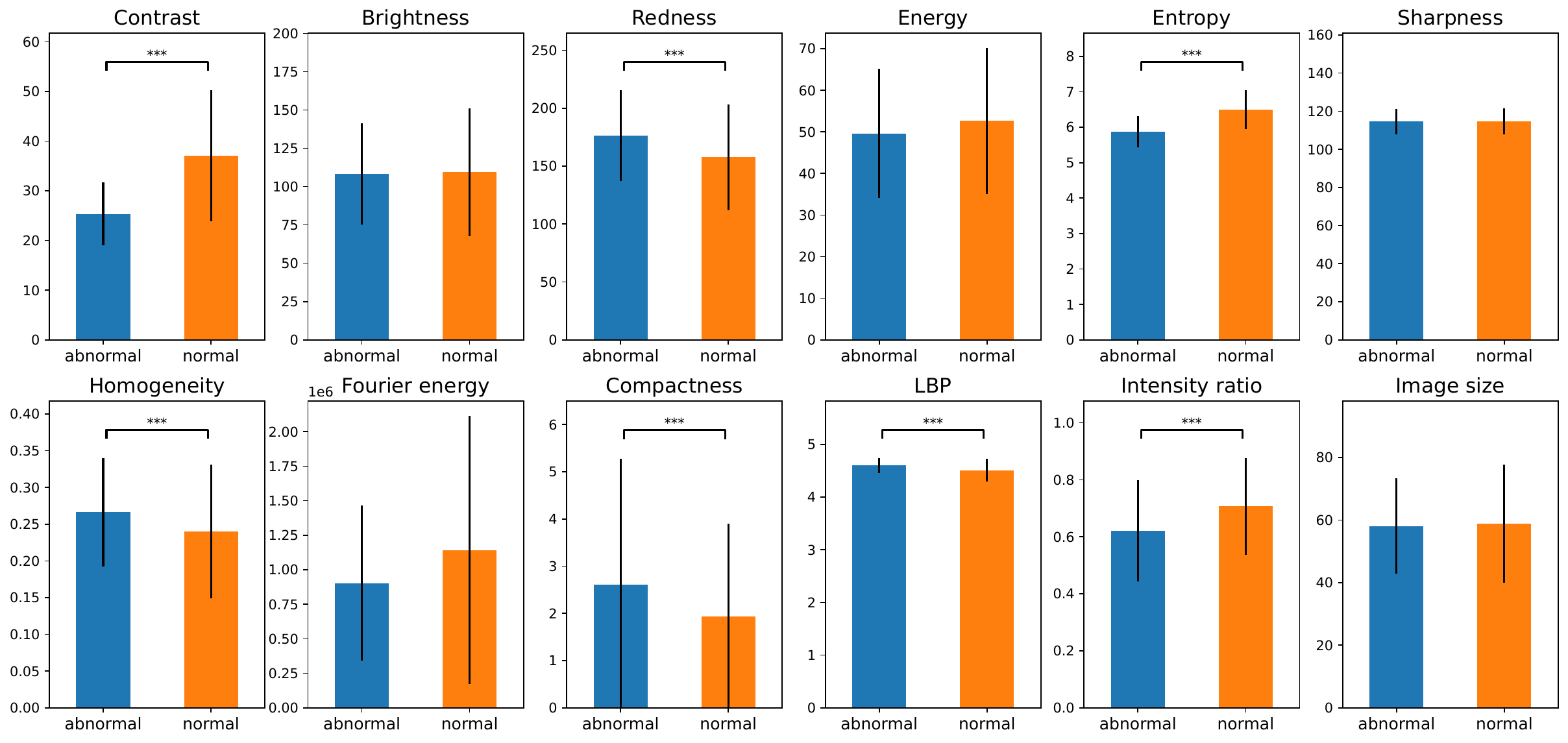}
    \caption{Image properties versus label for the entire pupils dataset. Significant differences between the normal and abnormal images are indicated by three asterisks for \(p<0.001\).}
    \label{fig:properties_vs_label}
\end{figure*}

\section{Methodology} \label{sec:methodology}

\subsection{Image dataset}

\subsubsection{Data collection}

Images of 2,991 children and teenagers between 5 to 18 years of age were collected in elementary and high schools across Washington State, United States, and Mexicali, Mexico. The images were taken by an ophthalmologist during school vision check visits, always with the help of a school nurse, and with written informed consent from the children’s legal guardians, in accordance with institutional ethical guidelines. The images were taken with multiple smartphone devices using both the Apple (iOS) and Android operating systems. The setting for taking the images was standardized: the child was placed in a semi-dark room, and the camera was positioned 1~m from the child. When possible, the child was facing a dark wall so that the camera flash would illuminate through larger pupils. This setup was optimized for the appearance of a red-eye reflex. Each image was annotated by the expert for visual impairments (if any).

\subsubsection{Data curation and preprocessing}
Each full-face image was cropped into two separate eyes using Google Cloud Vision's \cite{GoogleCloudVision} detection feature. The two eyes were treated as independent data points, thus extending the total dataset size to 5,982 individual images. For consistency, the left eye images were mirrored. In each of these eye images, a MMDetection model \cite{chen2019mmdetectionopenmmlabdetection} was used to localize and crop the pupil. To determine whether the image was usable, the light reflex was detected from the brightest spots in the pupil, as captured with a whiteness score. Then a hysteresis threshold selected the most intense reflex regions, with an upper threshold at the maximum whitness score and a lower threshold one standard deviation away. In case of multiple reflexes, the reflex closest to the center of the pupil was selected. The dataset was filtered according to these reflection sizes: if the reflection size was too big, too small, or too elongated, then the images were determined as unusable. The images were labeled according to the ophthalmologist's reports as normal, i.e. no visual impairments found, and abnormal, i.e. at least one visual impairment found. The final dataset contains 2,403 images, of which 1,728 were labeled as normal and 675 were labeled as abnormal. Finally, the images were split into training, validation, and testing datasets, with proportions of \(50\%\), \(25\%\), and \(25\%\), respectively. The dataset is not publicly available due to ethical and privacy considerations, but access may be granted upon reasonable request.

\subsection{Analysis of image properties} \label{subsec:properties}

The cropped pupils were analyzed for differences between the normal and abnormal images according to 12 different image properties (see below). We initially explored the use of a support vector machine (SVM) classifier to detect visual impairments directly from these properties. The SVM results (not shown) were poor and unreliable, thus encouraging the transition to deep neural networks. Nevertheless, the analysis of the image properties is insightful. The image properties chosen were the following:

\begin{itemize}
\item\textbf{Contrast}: Represents the amount of edges and sharp features, captured by the standard deviation of the grayscale image.
\item\textbf{Brightness}: Represents overall lightness, measured by the average pixel intensity of the grayscale image.
\item\textbf{Redness}: Computes the intensity of the red shades in the image by taking the average pixel intensity of the red channel. This is at the basis of the Bruckner test.
\item\textbf{Energy}: Measures the energy of the image, by convoluting over the grayscale image with a Laplacian kernel. High energy indicates more texture and variability.
\item\textbf{Entropy}: Measures the disorder in the image, computed by the Shannon entropy, indicating the amount of information and the complexity of an image.
\item\textbf{Sharpness}: Indicates the amount of fine details and sharp edges of an image. It is calculated as the variance of a Laplacian-filtered grayscale image.
\item\textbf{Homogeneity}: Measures how uniform the pixel intensities are by calculating the homogeneity of the grayscaled co-occurrence matrix (GLCM).
\item\textbf{Fourier Energy}: Identifies patterns in the spatial frequencies of an image, by transforming the grayscale image into Fourier space and adding the absolute values of the shifted Fourier transform.
\item\textbf{Compactness}: Measures the density of an image, calculated by the ratio of the area of a region to the area of a circle with the same perimeter.
\item\textbf{LBP (Local Binary Patterns)}: Captures local patterns of a grayscale image by comparing the intensity of a central pixel with its surrounding pixels in an area.
\item\textbf{Intensity Ratio}: Measures the intensity differences inside of an image by dividing the maximum intensity of the grayscale image by its minimum.
\item\textbf{Image Size}: Represents the original dimension of the image (width and height) in pixels.
\end{itemize}

\subsection{Deep Neural Networks}

\subsubsection{Models}

The deep neural network models chosen for this study were some of the most popular (relatively) lightweight architectures in the field of image classification: the ResNet-18 and the ResNet-34, the DenseNet-121, the EfficientNet-B0, the ConvNeXt-Tiny, and the SwinTransformer-Tiny. All these models were taken from the Torchvision package \cite{torchvision2016} for transfer learning from their corresponding weights pre-trained on ImageNet. These weights were frozen for our experiments, with the exception of the batch normalization parameters. For each model, the classification heads (last layers) were removed and replaced with a new 2-layer classifier that maps the flattened output features to a hidden layer of 512 units and then to an output layer of 2 units, for the normal and abnormal classes. The choice for this non-linear classifier head rather than a more conventional linear classifier was informed by an exploration of the latent space at the output of the pre-trained models.

\begin{table*}[!h]
\caption{Classification statistics for the different neural networks trained in this work.}
\label{tab:models}
\begin{center}
\begin{tabular}{|l|c|c|c|c|c|c|c|c|}
\hline
\textbf{Model} & \textbf{Parameters} & \textbf{Precision} & \textbf{Recall} & \textbf{Specificity} & \textbf{Accuracy} & \textbf{F1-Score} & \textbf{ROC-AUC}\\
\hline
\textbf{ResNet-18}            & 11.7M & \(0.65\pm 0.04\)          & \(0.82\pm0.05\)  & \(0.89\pm0.03\)          & \(0.88\pm0.02\)   & \(0.72\pm0.02\) & \(0.94\pm0.01\) \\
\textbf{ResNet-34}            & 21.8M & \(0.63\pm 0.06\)          & \(0.82\pm0.04\)  & \(0.88\pm0.04\)          & \(0.87\pm0.03\)   & \(0.71\pm0.04\) & \(0.94\pm0.01\) \\
\textbf{DenseNet-121}         & 8.0M  & \(0.53\pm 0.05\)          & \(0.71\pm0.05\)  & \(0.85\pm0.04\)          & \(0.82\pm0.02\)   & \(0.61\pm0.01\) & \(0.88\pm0.00\) \\
\textbf{EfficientNet-B0}      & 5.3M  & \(0.56\pm 0.05\)          & \(0.77\pm0.05\)  & \(0.85\pm0.04\)          & \(0.84\pm0.02\)   & \(0.64\pm0.02\) & \(0.90\pm0.01\) \\
\textbf{ConvNext-Tiny}        & 28.6M & \(0.57\pm 0.05\)          & \(0.76\pm0.05\)  & \(0.86\pm0.03\)          & \(0.85\pm0.02\)   & \(0.65\pm0.02\) & \(0.91\pm0.00\) \\
\textbf{SwinTransformer-Tiny} & 87.8M & \(\mathbf{0.71\pm 0.05}\) & \(0.78\pm0.04\)  & \(\mathbf{0.92\pm0.02}\) & \(0.89\pm0.01\)   & \(0.74\pm0.01\) & \(0.95\pm0.00\) \\
\hline
\textbf{Ensemble (all)}       & --   & \(0.68\)                  & \(0.82\)         & \(0.91\)                 & \(0.89\)          & \(0.74\)                   & -- \\
\textbf{Ensemble (best*)}     & --   & \(0.70\)                  & \(\mathbf{0.84}\)& \(0.91\)                 & \(\mathbf{0.90}\) & \(\mathbf{0.76}\)          & \(\mathbf{0.96 \pm 0.01}\) \\
\hline
\end{tabular}
\end{center}
\vspace*{-0.1cm}
\footnotesize{\hspace{1.2cm}* Ensemble of ResNet-18, ResNet-34, and SwinTransformer}
\end{table*}

\begin{table}[!b]
\caption{Classification statistics for different image augmentations used to train a ResNet-18 model.}
\label{tab:augmentations}
\begin{center}
\begin{tabular}{|l|c|c|c|}
\hline
\textbf{Augmentation} & \textbf{Accuracy} & \textbf{F1-Score} & \textbf{ROC-AUC}\\
\hline
\textbf{None}         & \(\mathbf{0.88\pm0.02}\) & \(0.72\pm0.02\) & \(0.94\pm0.01\) \\
\textbf{Color jitter} & \(0.87\pm0.02\)          & \(0.72\pm0.02\) & \(0.95\pm0.01\) \\
\textbf{Gaussian blur}     & \(0.87\pm0.03\)          & \(0.72\pm0.04\) & \(0.94\pm0.01\) \\
\textbf{Equalize}     & \(0.86\pm0.03\)          & \(0.71\pm0.05\) & \(0.93\pm0.01\) \\
\textbf{Contrast}     & \(0.86\pm0.03\)          & \(0.71\pm0.04\) & \(0.93\pm0.01\) \\
\textbf{Sharpness}    & \(0.86\pm0.01\)          & \(0.70\pm0.01\) & \(0.93\pm0.01\) \\
\textbf{Mirroring}    & \(0.86\pm0.02\)          & \(0.71\pm0.03\) & \(0.94\pm0.01\) \\
\textbf{Translation}  & \(0.87\pm0.02\)          & \(0.72\pm0.03\) & \(0.95\pm0.01\) \\
\textbf{Perspective}  & \(0.87\pm0.04\)          & \(0.72\pm0.04\) & \(0.94\pm0.01\) \\
\textbf{Rotation}     & \(\mathbf{0.88\pm0.03}\) & \(0.73\pm0.03\) & \(\mathbf{0.96\pm0.01}\) \\
\hline
\textbf{Mix (all)}     & \(\mathbf{0.88\pm0.02}\) & \(0.73\pm0.03\) & \(0.94\pm0.01\) \\
\textbf{Mix (best*)}     & \(\mathbf{0.88\pm0.03}\) & \(\mathbf{0.74\pm0.04}\) & \(\mathbf{0.96\pm0.00}\) \\
\hline
\multicolumn{4}{l}{
\vspace*{-0.01cm}
\footnotesize{* Mixture of Color jitter, Equalize, Sharpness, Translation,}
} \\
\multicolumn{4}{l}{
\vspace*{-0.05cm}
\footnotesize{\hspace*{.25cm}Perspective, and Rotation.}
} \\
\end{tabular}
\end{center}
\end{table}

\subsubsection{Training}

All models were trained on the training images, previously resized to \(224\times224\) pixels to fit the input size of the neural networks. We used a batch size of 64 images, a cross-entropy loss, and the AdamW optimizer with a learning rate of 0.001 and a weight decay of 0.01. Training continued for a maximum of 50 epochs, and the model with the lowest validation loss was saved. The entire procedure was repeated for 10 random seeds per model, with the same pre-trained weights but randomly initialized classifiers.

\subsubsection{Evaluation}

After training, the models were frozen and evaluated with the unseen test dataset. Their outputs were compared with the labels provided by the experts to compute the relevant summary statistics: precision, recall, specificity, accuracy, and F1-score. Additionally, we computed the classifier's prediction probabilities using a softmax activation of the output values. Using these probabilities, we obtained the ROC curves, with their corresponding area under the curve (ROC-AUC) scores. We also computed the model's confidence as the prediction probability of the winner output node, between 0.5 and 1. Additional evaluation methods are described in the results.

\section{Results} \label{sec:results}

\subsection{Image properties}

The image properties evaluated on the entire pupils dataset, i.e. combining training, validation, and testing data, are presented in Fig.~\ref{fig:properties_vs_label}. It is immediate to see that some of the properties have notorious differences between the normal and abnormal classess. In particular, the contrast of the abnormal images is lower than that of the normal images, whereas their compactness and homogeneity are higher. To further understand the results, we performed a Two-sample Kolmogorov-Smirnov (KS) significance test. Many of the image properties have significant differences. The redness of the image, which measures the average value of the red channel, is higher for the abnormal class. Other properties with significant differences, such as entropy and LBP, can provide hints about the structure underlying the images. However, as mentioned previously, these properties alone are not sufficient to correctly classify the images as normal or abnormal. 

\subsection{Models comparison}

The classification statistics of the trained models are shown in Table~\ref{tab:models}. The reported values are averaged over the 10 random seeds. Overall, we find excellent classification scores, with accuracies between 0.8 and 0.9 and recalls between 0.7 and 0.8 for all models. The individual model that achieves the best results is the SwinTransformer. However, it should be noted that this model has nearly 90 million parameters. On the other end of the spectrum, the EfficientNet-B0 only has 5 million parameters but achieves a comparable performance. The ResNet-18 and ResNet-34, two very popular models in image analysis, achieve the highest recall score, 0.82, while also having accuracies close to 0.9. The ResNets and the SwinTransformer also share the highest ROC-AUC scores.

We create two ensemble models that combine the outcomes of (i) all models and (ii) the best models, i.e. the SwinTransformer, the ResNet-18, and the ResNet-34. The latter achieves the highest scores overall.

\begin{figure*}[!t]
    \centering
    \begin{minipage}{0.48\textwidth}
        \centering
        \includegraphics[width=0.95\textwidth]{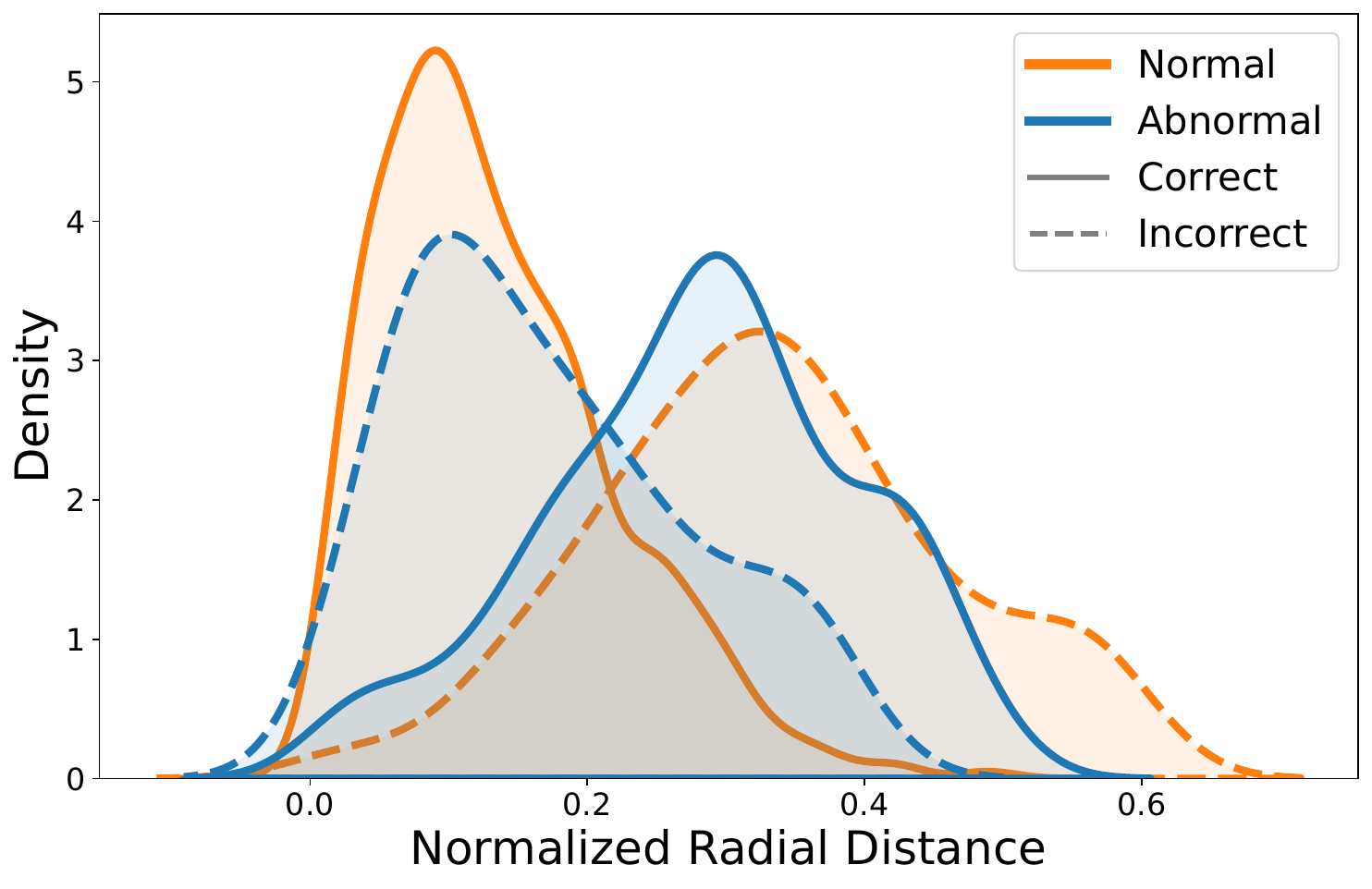}
        \caption{Radial distributions of attention maps focus.}
        \label{fig:attention-distribution}
    \end{minipage}
    \hfill
    \begin{minipage}{0.48\textwidth}
        \centering
        \begin{subfigure}[t]{0.48\textwidth}
            \centering
            \includegraphics[width=\textwidth]{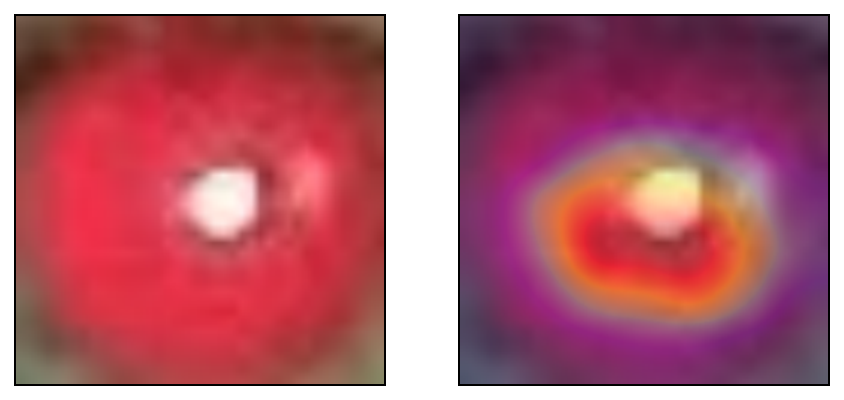}
            \caption{Normal, correct}
        \end{subfigure}
        \hfill
        \begin{subfigure}[t]{0.48\textwidth}
            \centering
            \includegraphics[width=\textwidth]{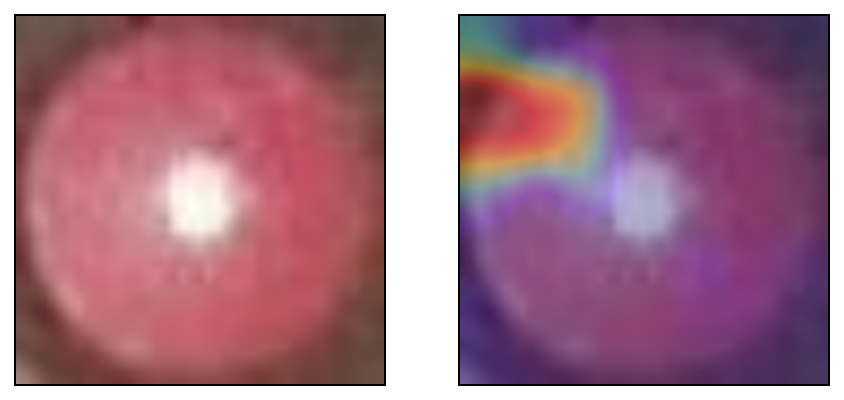}
            \caption{Normal, incorrect}
        \end{subfigure}
        \\[1em]
        \begin{subfigure}[t]{0.48\textwidth}
            \centering
            \includegraphics[width=\textwidth]{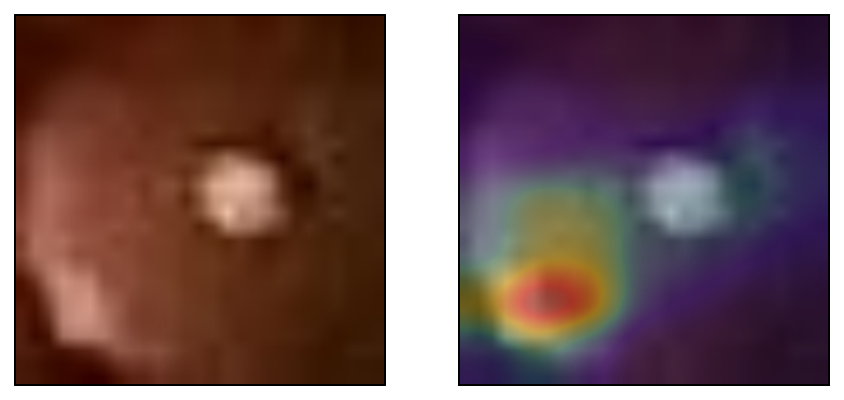}
            \caption{Abnormal, correct}
        \end{subfigure}
        \hfill
        \begin{subfigure}[t]{0.48\textwidth}
            \centering
            \includegraphics[width=\textwidth]{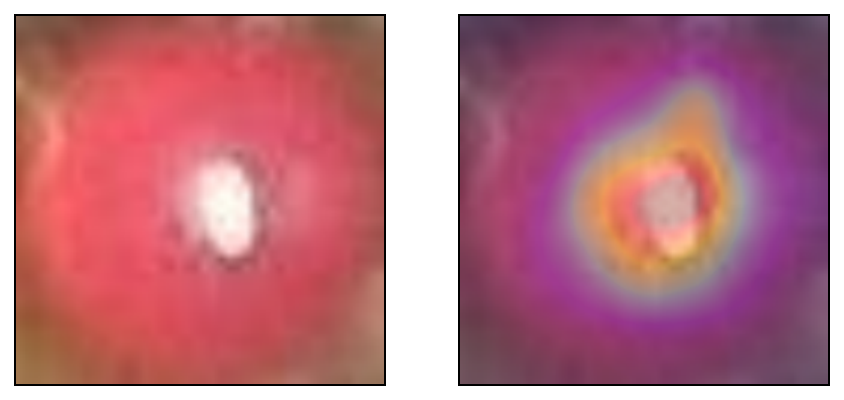}
            \caption{Abnormal, incorrect}
        \end{subfigure}
        \caption{Examples of attention maps from the test dataset.}
        \label{fig:attention-examples}
    \end{minipage}
    \label{fig:attention-combo}
\end{figure*}

\subsection{Image augmentations comparison}

To address the dataset's imbalance in the number of normal and abnormal images, we explore the use of image augmentations. These can artificially increase the variability in abnormal images, making the classifier more robust. We repeat 10 additional training iterations of the ResNet-18 for each of the augmentations mentioned in Table~\ref{tab:augmentations}, with the default values from the Torchvision package. The results shown reveal that some augmentations have a positive impact. We also train models with mixtures of augmentations, and we find that the best results are obtained with a mixture of the following: color jitter, equalize, sharpness, translation, perspective, and rotation.

\subsection{Model interpretability}

The trained models achieve excellent classification. However, since this work is the basis for a medical application, it is crucial to ensure that the classifier is trustworthy. Here, we aim to gain insights about how the model processes images that can allow us to provide confident feedback to future users. All the following results are obtained from one particular training iteration of a ResNet-18 model.

\subsubsection{Attention maps}

First, we visualize the regions that the model is using to make decisions. To this end, we use Grad-CAM \cite{selvaraju2017grad} to create an attention map based on the gradient computed at the last convolutional layer of the ResNet. Some examples are shown in Fig.~\ref{fig:attention-examples}. Additionally, we compute the radial distance of the most attended region of each image. In Fig.~\ref{fig:attention-distribution}, we plot the distributions of the normalized attention radial distances for the normal and abnormal classes. We find that the attention maps of normal images focus on the center of the image, but the opposite is true for abnormal images. However, incorrectly classified images display an inverted attention behavior. This suggests that misclassification is associated with attending to spatially inappropriate regions. This result highlights a strong link between the model's attention localization and its decision outputs.

\subsubsection{Latent space}

A common approach to understand and explain how the models perform is to visualize the latent space, i.e. the high-dimensional vectors that the networks use to drive the classification. In Fig.~\ref{fig:tsne}, we plot the distribution of test images at the output level of the ResNet-18, using the t-SNE dimensionality reduction algorithm \cite{van2008visualizing}. While the normal and abnormal images are mostly clustered separately, this visualization reveals that the incorrectly classified images lie in a region around the decision boundary that separates the two classes. This t-SNE plot can be used to evaluate new unlabeled images: ideally, they should lie as far as possible from the decision boundary. If this is not the case, the model's output may not be reliable.

\subsubsection{Confidence and image properties}

\begin{figure*}[!t]
    \centering
    \begin{minipage}[t]{0.32\textwidth}
        \centering
        \includegraphics[width=0.95\textwidth]{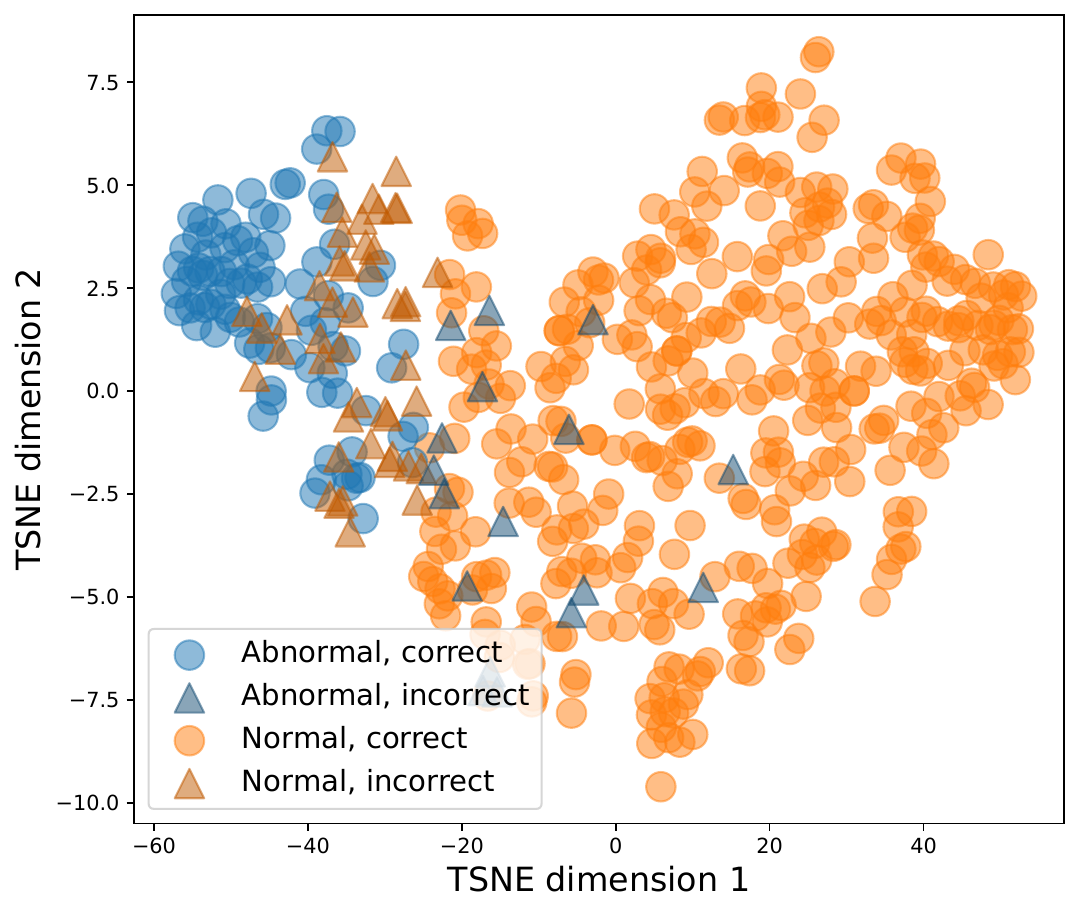}
        \caption{Latent space t-SNE visualization.}
        \label{fig:tsne}
    \end{minipage}
    \hfill
    \begin{minipage}[t]{0.64\textwidth}
        \centering
        \includegraphics[width=0.99\textwidth]{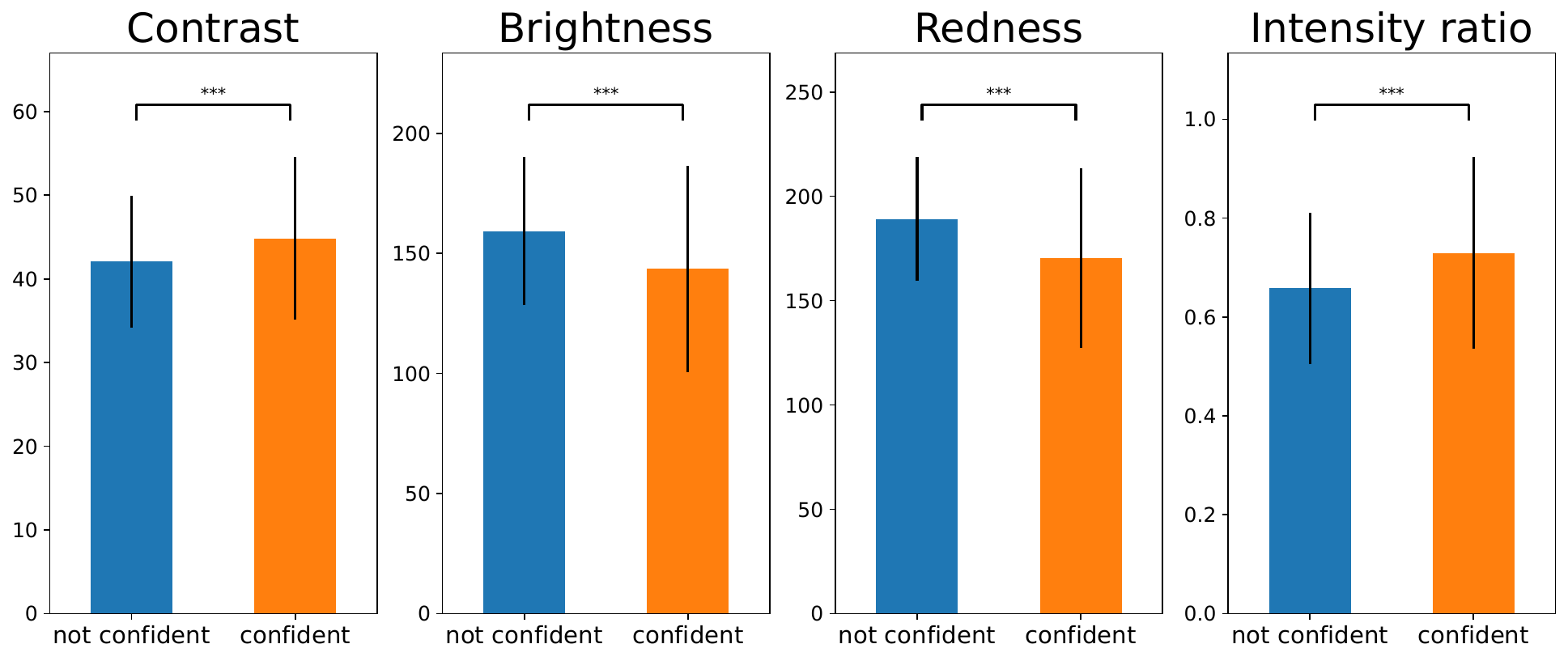}
        \caption{Image properties with significant differences against model confidence.}
        \label{fig:properties_vs_confidence}
    \end{minipage}
\end{figure*}

Building on the latent space analysis, we inquire whether it is possible to provide immediate feedback to users so that they can provide a new (more reliable) image. To do so, we circle back to the image properties detailed in Section \ref{sec:methodology}. We collect the model confidences for the test images and split the dataset in half relative to the median, resulting in two groups: confident and not confident. In parallel, we compute the image properties on the full-eye images extracted by Google Vision, i.e. before cropping the pupils (see Fig.~\ref{fig:KVC_pipeline}). Only four properties have significant differences following a Two-sample Kolmogorov-Smirnov (KS) test, shown in Fig.~\ref{fig:properties_vs_confidence}. Images with higher classification confidence have higher contrast, lower brightness, lower redness, and higher intensity ratios. That is, dark backgrounds and bright reflexes of the retina should be preferred. With this information, we can request users of the KidsVisionCheck app to take new photos whenever their pupil images lie too close to the decision boundary or when the model is not confident in its classification, potentially reducing misclassifications.



\section{Discussion} \label{sec:discussion}

To summarize, we find that the best classification performance is achieved by an ensemble of models (ResNet-18, ResNet-34, SwinTransformer-Tiny) and by the use of a mixture of image augmentations during training. The combination of both yields an accuracy of \(0.90\) with an F1-score of \(0.77\) and an ROC-AUC of \(0.96\). This model is in the deployment stage to be used in the KidsVisionCheck app.

Despite the positive outcomes, this study has a few notable limitations. The models exploit transfer learning from neural networks pretrained on ImageNet, which has a very different structure from the pupil dataset presented here. It remains to be seen whether the use of alternative pretrained backbones, or at least additional finetuning of the networks, could have a positive impact on the results.

Furthermore, a number of limitations were unveiled after the data collection was finalized. The number of images is relatively large compared with other similar datasets, but it is also  imbalanced. The labels were provided by a single ophthalmologist without any cross-validation, meaning that subjective biases might skew the classification. Finally, all the images used during training and testing were taken for the purpose of this study. During deployment, new images can be taken under out-of-distribution conditions, which could negatively impact performance.

The Grad-CAM results reveal interesting differences in the attended regions for normal and abnormal images. However, the corresponding ground truth is not available because the expert labeling was only categorical. Recent works have shown that incorporating such expert knowledge during training (e.g. from eye tracker recordings \cite{jiang2023eye}) can result in better model alignment and a boost in performance.

\section{Conclusions} \label{sec:conclusions}

This work presented the collection and preprocessing of a dataset of nearly 2,500 red-eye reflex pupil images from children. With the use of state-of-the-art machine learning models, we achieved a classification accuracy of 90\%, demonstrating the feasibility of detecting visual impairment abnormalities without an ophthalmologist. Furthermore, it was possible to obtain several insights about how the models perform this classification and which image properties are desired for confident classification. However, the dataset used for this work had many limitations. To address these, a new iteration of data collection is planned for the near future. Following the analysis presented here, we aim to guarantee higher quality, variability, and robustness. Detailed expert labels with fine-grained information about specific vision disorders will allow us to expand the model's diagnostic capabilities. The long term goal for KidsVisionCheck is to provide free, easily accessible, and accurate pediatric vision screenings and enable early intervention for vision abnormalities for children around the world.

\end{document}